\documentclass{article}
\usepackage{spconf,amsmath,graphicx,cite}
\usepackage{svg}
\DeclareMathOperator*{\argmax}{arg\,max}

\title{CLASS BASED THRESHOLDING IN EARLY EXIT SEMANTIC SEGMENTATION NETWORKS}
\name{Alperen Görmez and Erdem Koyuncu}
\address{Department of Electrical and Computer Engineering, University of Illinois Chicago, Chicago, IL}

\begin{document}

\maketitle

\begin{abstract}
We propose Class Based Thresholding (CBT) to reduce the computational cost of early exit semantic segmentation models while preserving the mean intersection over union (mIoU) performance. A key idea of CBT is to exploit the naturally-occurring neural collapse phenomenon. Specifically, by calculating the mean prediction probabilities of each class in the training set, CBT assigns different masking threshold values to each class, so that the computation can be terminated sooner for pixels belonging to easy-to-predict classes. We show the effectiveness of CBT on Cityscapes and ADE20K datasets. CBT can reduce the computational cost by $23\%$ compared to the previous state-of-the-art early exit models.
\end{abstract}


\section{Introduction}
Deep learning is developing fast, and new state-of-the-art models are continuously being announced. The increase in the performance of the state-of-the-art models is often due to the increased model size \cite{era, opt, gpt}. Larger models can learn more complex patterns and hence achieve higher performance, but they also require more floating point operations, which means they have higher inference cost. In an age where everything is being decentralized to run at the edge (e.g. mobile phones and IoT devices), it is important to reduce the inference cost of large models to be able to deploy them to resource-constrained devices.

\begin{figure}[htb]
  \centering
  \centerline{\includesvg[width=7.5cm]{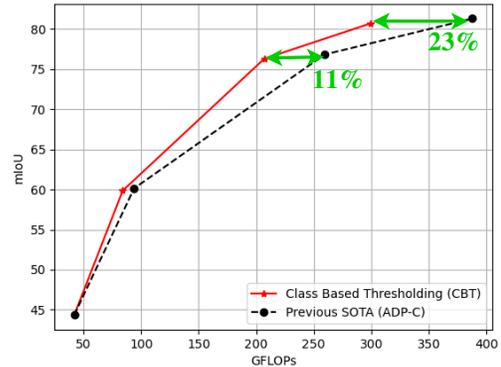}}
\vspace{-10pt}
\caption{Comparison of CBT with the previous state-of-the-art on the Cityscapes dataset for HRNetV2-W48 model.}
\vspace{-8pt}
\label{fig:fig1}
\end{figure}

In order to reduce the amount of computation without harming the model performance, various methods have been used. Knowledge distillation methods start from scratch and train a smaller model based on the output of the larger model \cite{kdistill}. Quantization methods reduce the bit length of the model weights \cite{quant_prune}. Pruning methods set redundant model weights to zero so that these weights are not used in the computations \cite{quant_prune, pruneee}. Early exit networks allows ``easy'' data samples to exit the model early to save computation \cite{e2cm, shallowdeep}. Among these methods, early exit networks exploit the fact that real world data is heterogeneous, i.e., not all data samples have the same ``difficulty''. Early exit networks have also close ties with the phenomenon of \emph{neural collapse} \cite{e2cm, neuralcollapse}.

The neural collapse phenomenon states that as one travels deeper in a neural network model, the intermediate representations of the data samples become more and more disentangled and distinct clusters can be identified at the last layer, which makes classification easier \cite{neuralcollapse}. Recent works expand on this phenomenon and show that clusters begin to form even at earlier layers \cite{cascadingcollapse,  e2cm}, resulting in a so-called \emph{cascading collapse}. In the supervised setting, each cluster corresponds to a class where the model is trained on, and the mean of the cluster is referred to as simply a \emph{class mean}. 

\begin{figure*}
  \centering
  \centerline{\includegraphics[width=0.79\linewidth]{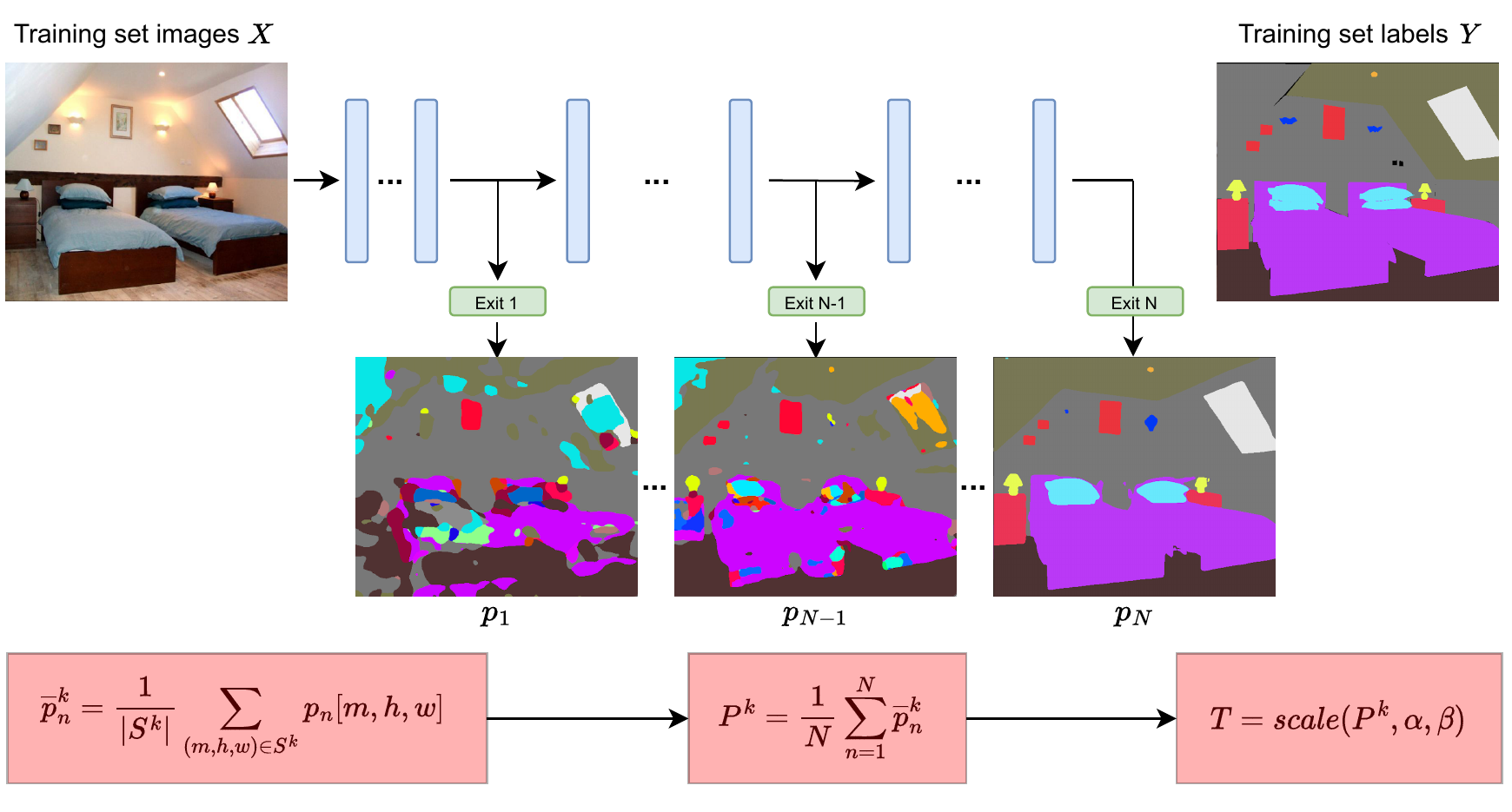}}
\vspace{-10pt}
\caption{Overview of Class Based Thresholding for $$N$$ exit layers and $$K$$ classes. Best viewed in color and zoom in.}
\vspace{-6pt}
\label{fig:fig2}
\end{figure*}

For the task of image classification, we have previously shown that one can design a low-complexity early exit network by taking advantage of the neural collapse phenomenon \cite{e2cm}. Specifically, a feature that is sufficiently close to a class mean at any given layer can be allowed an early exit, without significant penalty in classification performance. However, the same design is not immediately applicable to the task of semantic segmentation since one now needs to perform pixel-wise classification. In fact, in an image classification task, there is one input and it belongs to one class. Therefore the intermediate layer outputs will be close to only one class mean, and a meaningful prediction can be performed based on the distances to the class means \cite{e2cm}. On the contrary, in semantic segmentation, one input has many pixels that belong to many classes. Moreover, the spatial locations of the pixels matter. These make it infeasible to calculate the class means for the pixels using existing algorithms (e.g. \cite{e2cm}). Nevertheless, utilizing the neural collapse phenomenon for semantic segmentation would be particularly useful because state-of-the-art semantic segmentation models preserve high resolution intermediate representations throughout the model, which increases the amount of computation significantly \cite{hrnet, icassp20, icassp21, icassp22}.

In this work, we propose a novel algorithm, ``Class Based Thresholding (CBT)'', which reduces the computational cost while preserving the model performance for the semantic segmentation task. Similar to previous state-of-the-art, CBT employs a thresholding mechanism to allow the early termination of the computation for confidently predicted pixels. CBT utilizes the neural collapse phenomenon by calculating the mean of the prediction probabilities of pixels in the training set, for each class. The thresholds for each class are calculated via a simple transformation of the class means. We show the effectiveness of CBT on the Cityscapes \cite{cityscapes} and ADE20K \cite{ade20k} datasets using the HRNetV2-W18 and HRNetV2-W48 models \cite{hrnet}. By efficiently utilizing the neural collapse phenomenon, CBT can reduce the computational cost by up to $23\%$ compared to the previous state-of-the-art method while preserving the model performance as shown in Fig.~\ref{fig:fig1}.

\section{CLASS BASED THRESHOLDING}
We build on the state-of-the-art early exit semantic segmentation method, ``Anytime Dense Prediction with Confidence Adaptivity (ADP-C)'' \cite{adpc}. ADP-C adds early exit layers to the base semantic segmentation model and introduces a masking mechanism based on a single user-specified threshold value $t$ to reduce the computational cost. If a pixel is predicted confidently at an exit layer, i.e., the maximum prediction probability over all classes is greater than the threshold $t$, that pixel is  masked for all subsequent layers. Any masked pixel will not be processed again at later layers. The computational cost is reduced due to the induced feature sparsity.

A big room for improvement for ADP-C stems from the observation that the same user-specified threshold value $t$ is used for every class. However, it is more plausible that different threshold values should be used for different classes, and the threshold values should reflect the dataset and class properties, rather than just being a user-specified number. This is because pixels belonging to different classes have different difficulty levels of being predicted correctly. For example, using $t=0.998$ for \emph{person} class as in ADP-C makes sense because we may want to be really certain about pixels belonging to people. However, pixels belonging to the \emph{sky} class will be often easier to predict than pixels belonging to the \emph{person} class, which means the model will be confident about them much sooner. Therefore, a lower threshold value can be used for the \emph{sky} class without significant penalty in prediction accuracy. Otherwise, more computation will have to be performed for the \emph{sky} pixels although it is not necessary.

Given a model trained on a semantic segmentation task with $K$  classes, we propose using different masking threshold values for each class, based on the dataset and class properties. Let $T = [T_1 \cdots T_K] \in {[0,1]}^K$ be the threshold vector that we wish to determine, where the $k^{th}$ element $T_k$ corresponds to class $k$, and ${k \in \{1,2,\ldots,K\}}$. Let there be $N$ exit layers in the model. Let $p_n$ denote the prediction probabilities at layer $n$, where ${n \in \{1,2,\ldots,N\}}$. 

Our algorithm is illustrated in Fig.~\ref{fig:fig2}. At each exit layer ${n \in \{1,2,\ldots,N\}}$, for each class ${k \in \{1,2,\ldots,K\}}$ in the training set, we calculate the mean of layer $n$'s prediction probabilities using all training set pixels that belong to class $k$ (denoted as $S^{k}$). This yields 
\begin{equation}
\overline{p}_n^{k} = {\frac {1}{|{S}^k|}}\sum _{(m,h,w) \in S^{k}}{p}_n[m,h,w] \in {[0,1]}^K.
\label{eq1}
\end{equation} 
The $i^{th}$ element of $\overline{p}_n^{k}$ denotes the average probability of a pixel belonging to class $i$ when the ground truth for that pixel is class $k$. Next, we compute 
\begin{equation}
{P}^k = {\frac {1}{N}}\sum _{n=1}^{N}\overline{p}_n^{k},
\label{eq2}
\end{equation} 
which is the average of $\overline{p}_n^{k}$ over all layers. Hence, information across layers is shared.

We initialize the threshold $T_k$ to be the difference between the largest and the second largest elements of $P^k$. The difference serves as a confidence score. If the confidence score is high, then the masking threshold should be low so that the computation can terminate easily. After all components of $T$ are initialized in this manner, we inversely scale $T$ according to two parameters $\alpha$ and $\beta$ so that the maximum and minimum class confidence scores determined by $T$ will be converted to masking threshold values $\alpha$ and $\beta$ respectively, where $\alpha<\beta$. Specifically, the scaling is done via a single application of the update rule

\begin{equation}
T_k \leftarrow \left(1 - \frac{T_k - \min_{\ell} T_{\ell}}{\max_{\ell} T_{\ell} - \min_{\ell} T_{\ell}}\right)(\beta-\alpha) + \alpha.
\label{eq3}
\end{equation}

The inference is performed as follows: Let $\pi \in {[0,1]}^K$ be the prediction probabilities for a pixel at an exit layer. Let $j = \argmax \pi$. If $\pi_j > T_j$, this pixel will be marked as confidently predicted (predicted as class $j$) and will be incorporated to the mask $M$ as in ADP-C \cite{adpc}. The mask will have $0$ at the locations of the confidently predicted pixels, and $1$ at every other place. By doing so, the outputs of subsequent layers at these locations will not be calculated. Instead, already computed values will be used.

\def\arraystretch{1.2}
\begin{table*}[t]
\caption{Results on Cityscapes.}
\label{tab:table1}
\vspace{-4pt}
\begin{center}
\begin{tabular}{|c|c|c|c|c|c|c|c|c|c|}
\hline
& & \multicolumn{8}{c|}{ \textbf{Exit}} \\
\cline{3-10}
 \textbf{Method} &  \textbf{Model} & \multicolumn{2}{c|}{ \textbf{1}} & \multicolumn{2}{c|}{ \textbf{2}} & \multicolumn{2}{c|}{ \textbf{3}} & \multicolumn{2}{c|}{ \textbf{4}}\\
\cline{3-10}
& & \textbf{mIoU} & \textbf{GFLOPs} & \textbf{mIoU} & \textbf{GFLOPs} & \textbf{mIoU} & \textbf{GFLOPs} & \textbf{mIoU} & \textbf{GFLOPs} \\ \hline
ADP-C & HRNetV2-W48 & 44.34 & 41.92 & 60.13 & 93.90 & 76.82 & 259.33 & 81.31 & 387.80 \\ \hline
CBT [0.99, 0.998] & HRNetV2-W48 & 44.34 & 41.92 & 59.85 & 84.02 & 76.29 & 206.89 & 80.69 & 299.10 \\ \hline
CBT [0.95, 0.998] & HRNetV2-W48 & 44.34 & 41.92 & 57.97 & 71.57 & 72.86 & 155.77 & 76.60 & 222.65 \\ \hline
CBT [0.90, 0.998] & HRNetV2-W48 & 44.34 & 41.92 & 56.05 & 65.91 & 68.92 & 132.49 & 72.29 & 186.31 \\ \hline
ADP-C & HRNetV2-W18 & 40.83 & 23.68 & 48.19 & 33.27 & 68.26 & 45.40 & 77.02 & 58.90 \\ \hline
CBT [0.99, 0.998] & HRNetV2-W18 & 40.83 & 23.68 & 48.07 & 31.74 & 67.98 & 41.40 & 76.57 & 51.26 \\ \hline
CBT [0.95, 0.998] & HRNetV2-W18 & 40.83 & 23.68 & 46.97 & 29.51 & 64.88 & 36.25 & 72.18 & 43.35 \\ \hline
CBT [0.90, 0.998] & HRNetV2-W18 & 40.83 & 23.68 & 45.79 & 28.39 & 61.32 & 33.72 & 67.45 & 39.42 \\ \hline
\end{tabular}
\end{center}

\vspace{-8pt}

\caption{Results on ADE20K.}
\label{tab:table2}
\vspace{-14pt}
\begin{center}
\begin{tabular}{|c|c|c|c|c|c|c|c|c|c|}
\hline
& & \multicolumn{8}{c|}{ \textbf{Exit}} \\
\cline{3-10}
 \textbf{Method} &  \textbf{Model} & \multicolumn{2}{c|}{ \textbf{1}} & \multicolumn{2}{c|}{ \textbf{2}} & \multicolumn{2}{c|}{ \textbf{3}} & \multicolumn{2}{c|}{ \textbf{4}}\\
\cline{3-10}
& & \textbf{mIoU} & \textbf{GFLOPs} & \textbf{mIoU} & \textbf{GFLOPs} & \textbf{mIoU} & \textbf{GFLOPs} & \textbf{mIoU} & \textbf{GFLOPs} \\ \hline
ADP-C & HRNetV2-W48 & 4.12 & 6.20 & 5.16 & 15.42 & 12.15 & 52.47 & 42.82 & 100.28 \\ \hline
CBT [0.90, 0.998] & HRNetV2-W48 & 4.12 & 6.20 & 5.15 & 15.07 & 12.09 & 50.48 & 41.85 & 94.31 \\ \hline
CBT [0.80, 0.998] & HRNetV2-W48 & 4.12 & 6.20 & 5.14 & 14.80 & 11.90 & 48.81 & 40.17 & 90.25 \\ \hline
CBT [0.70, 0.998] & HRNetV2-W48 & 4.12 & 6.20 & 5.12 & 14.55 & 11.58 & 47.27 & 37.54 & 86.52 \\ \hline
ADP-C & HRNetV2-W18 & 4.89 & 5.88 & 6.83 & 7.84 & 8.94 & 12.73 & 9.74 & 19.04 \\ \hline
CBT [0.90, 0.998] & HRNetV2-W18 & 4.89 & 5.88 & 6.80 & 7.73 & 10.07 & 12.24 & 11.78 & 17.89 \\ \hline
CBT [0.80, 0.998] & HRNetV2-W18 & 4.89 & 5.88 & 6.75 & 7.67 & 10.17 & 11.98 & 11.95 & 17.26 \\ \hline
CBT [0.70, 0.998] & HRNetV2-W18 & 4.89 & 5.88 & 6.70 & 7.62 & 10.09 & 11.75 & 11.88 & 16.71 \\ \hline
\end{tabular}
\end{center}
\vspace{-9pt}
\end{table*}

\section{RESULTS}
We validate the effectiveness of our method on Cityscapes \cite{cityscapes} and ADE20K \cite{ade20k} datasets using the HRNetV2-W18 and HRNetV2-W48 models \cite{hrnet}. We use mean intersection over union (mIoU) as our performance metric and number of floating point operations (FLOPs) as our computational cost metric. We report the performance on the validation sets.

\begin{figure}[htb]
  \centering
  \vspace{-16pt}
  \centerline{\includesvg[width=7.7cm]{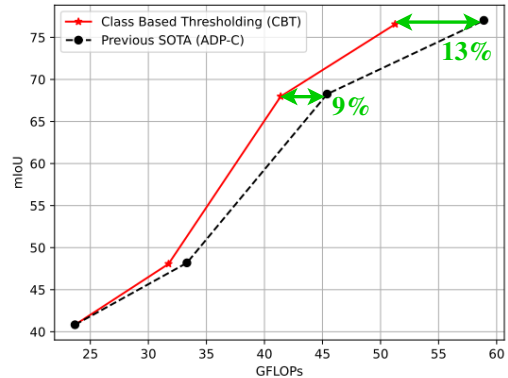}}
\vspace{-10pt}
\caption{Comparison of CBT with the previous state-of-the-art on the Cityscapes dataset for HRNetV2-W18 model.}
\vspace{-8pt}
\label{fig:fig3}
\end{figure}

\subsection{Datasets}
Cityscapes dataset consists of $2048\times1024$ RGB urban street images from various cities. There are $19$ classes in the dataset.

ADE20K dataset consists of RGB images of various scenes and objects. There are $20210$ training images and $2000$ validation images. The image sizes vary. There are $150$ classes in the dataset.

\subsection{Models}
HRNet is the state-of-the-art model for the semantic segmentation task. The intermediate outputs are kept in four different resolutions throughout the model \cite{hrnet}. We use HRNetV2 in our experiments, where the different resolutions are combined before giving an output. Similar to ADP-C, we use HRNetV2-W18 and HRNetV2-W48 in our experiments, where HRNetV2-W18 is more lightweight than HRNetV2-W48 due to smaller number of channels being used.

We attach $3$ early exit layers to HRNetV2-W18 and HRNetV2-W48 models as in ADP-C. The exit layer structures and positions are exactly the same as ADP-C. Each model has $N=4$ exits in total. We used the pretrained models provided by the authors of ADP-C for our Cityscapes experiments. For the ADE20K experiments, we trained the models ourselves. The training is done by using the weighted sum of the exit losses. Similar to ADP-C, we gave all exit losses the same weight of $1$. We used a single NVIDIA RTX A6000 GPU for training. We trained the models for $200000$ iterations. We used a batch size of $16$.

\subsection{Experiments}
We have evaluated CBT with numerous $\alpha$-$\beta$ pairs, and compared it with the baseline ADP-C \cite{adpc}. We kept $\beta=0.998$ in all our experiments for a fair comparison because ADP-C achieves the best performance with $t=0.998$ as stated in \cite{adpc}. As shown in Tables \ref{tab:table1} and  \ref{tab:table2}, we have also used a limited and uniform set of values for $\alpha$; namely $\alpha\in\{0.7,0.8,0.9,0.9,0.95,0.99\}$ so as to avoid excessive hyperparameter tuning.

The mIoU and GFLOPs for the first exit do not differ between CBT and ADP-C, because the masking procedure starts there. Until the first exit, the mask consists of all $1$'s so that all pixels are operated on. According to the prediction probabilities at Exit 1, the mask changes for deeper layers.

Looking at Table \ref{tab:table1}, it can be seen that CBT $[0.99, 0.998]$ decreases the computational cost by $23\%$ while losing only $0.62$ mIoU for HRNetV2-W48. For Exits 2 and  3, the computational cost is decreased by $10\%$ and $20\%$ respectively. By using smaller $\alpha$, the computational cost can be decreased more, but mIoU starts degrading as well. Also, decreasing $\alpha$ can result in worse performance per FLOP as can be seen from Exit 3 of CBT $[0.95, 0.998]$ and Exit 4 of CBT $[0.90, 0.998]$. It is also worth noting that Exit 4 of CBT $[0.95, 0.998]$ can match the performance of Exit 3 of ADP-C while using $14\%$ less computation. For HRNetV2-W18, the results follow the same trend as in HRNetV2-W48 on Cityscapes dataset. CBT $[0.99, 0.998]$ decreases the computational cost by $9\%$ and $13\%$ for Exit 3 and Exit 4 respectively as seen in Fig.~\ref{fig:fig3}.

From Table \ref{tab:table2}, we see that CBT can save computational cost on ADE20K dataset too, which has significantly more number of classes compared to Cityscapes. More specifically, CBT $[0.90, 0.998]$ decreases the computational cost by $6\%$ while losing only $0.97$ mIoU for HRNetV2-W48. The reason why the performances at the first three exit is low for both ADP-C and CBT is because the model cannot perform well enough due to large number of classes. It needs significantly more computation (e.g. $94.31$ GFLOPs instead of $15.07$) to have better performance. Also, this is why CBT cannot reduce the computational cost on ADE20K as much as it does on Cityscapes dataset. Another interesting observation is that when the the model size is small, CBT can increase the performance by up to $22\%$ compared to ADP-C even when $\alpha$ is as low as $0.7$.  We believe this is because when the model size is small and the dataset is complex, the model cannot reach high confidence for the majority of the pixels although it could predict them correctly. Relaxing the threshold helps model to perform correct predictions.

\section{CONCLUSION}
We proposed a novel algorithm that utilizes the naturally occurring neural collapse phenomenon to reduce the computational cost of early exit semantic segmentation models. Experiment results on different datasets and models suggest our method is effective in reducing the computational cost without significant penalty in model performance.

\bibliographystyle{IEEEbib}
\bibliography{refs}%

\end{document}